%% file: main.tex
\documentclass{article}

\usepackage{microtype}
\usepackage{graphicx}
\usepackage{subfigure}
\usepackage{booktabs} 
\usepackage[accepted]{mlsys2021}

\newcommand{\Fig}[1]{Figure~\ref{#1}}

\begin{document}

\twocolumn[
\mlsystitle{Exploring the limits of concurrency in ML Training on Google TPUs}

\begin{mlsysauthorlist}
\mlsysauthor{Sameer Kumar}{google}
\mlsysauthor{Yu Emma Wang}{google}
\mlsysauthor{Cliff Young}{google}
\mlsysauthor{James Bradbury}{google}
\mlsysauthor{Anselm Levskaya}{google}
\mlsysauthor{Blake Hechtman}{google}
\mlsysauthor{Dehao Chen}{google}
\mlsysauthor{HyoukJoong Lee}{google}
\mlsysauthor{Mehmet Deveci}{google}
\mlsysauthor{Naveen Kumar}{google}
\mlsysauthor{Pankaj Kanwar}{google}
\mlsysauthor{Shibo Wang}{google}
\mlsysauthor{Skye Wanderman-Milne}{google}
\mlsysauthor{Steve Lacy}{google}
\mlsysauthor{Tao Wang}{google}
\mlsysauthor{Tayo Oguntebi}{google}
\mlsysauthor{Yazhou Zu}{google}
\mlsysauthor{Yuanzhong Xu}{google}
\mlsysauthor{Andy Swing}{google}
\end{mlsysauthorlist}

\mlsysaffiliation{google}{Google Inc.}
\mlsyscorrespondingauthor{Sameer Kumar}{sameerkm@google.com}

\vskip 0.3in

\begin{abstract}
Recent results in language understanding using neural networks have required training hardware of unprecedented scale, with thousands of chips cooperating on a single training run. This paper presents techniques to scale ML models on the Google TPU Multipod, a mesh with 4096 TPU-v3 chips. We discuss model parallelism to overcome scaling limitations from the fixed batch size in data parallelism, communication/collective optimizations, distributed evaluation of training metrics, and host input processing scaling optimizations. These techniques are demonstrated in both the TensorFlow and JAX programming frameworks. We also present performance results from Google's recent submission to the MLPerf-v0.7 benchmark contest, achieving record-breaking training times from 16 to 28 seconds in four MLPerf models on the Google TPU-v3 Multipod machine.
\end{abstract}
]

\input{intro}
\input{frameworks}
\input{techniques}
\input{model}

\input{performance}
\input{summary}
\input{ack}
\bibliographystyle{mlsys2020}
\bibliography{main}

\end{document}

%% file: intro.tex
\section{Introduction}
The deep learning revolution is in the midst of a “space race” in the field of language understanding, with leading research labs training and publishing papers about a sequence of models of exponentially increasing size. One of the early breakthroughs was Google’s Neural Machine Translation System~\cite{wu2016google}, which used LSTMs~\cite{hochreiter1997long} and Attention~\cite{luong2014addressing,bahdanau2014neural} to achieve a significant quality improvement. GNMT was rapidly followed by Transformers-\cite{vaswani2017attention} which parallelized over input sequences, allowing faster training than sequentially limited LSTMs. Transformers in turn are a fundamental component of BERT~\cite{devlin2018bert} models, which are able to “pre-train” for general linguistic knowledge, then “fine-tune” to particular language tasks, including Translation. The latest GPT-3 model appears to be able to compose plausible essay-length arguments, albeit with some degree of human guidance or selection~\cite{brown2020language}. The size of these models is growing exponentially; OpenAI observed that the training resources for state-of-the-art deep learning models appears to be doubling every 3.5 months~\cite{aiandcompute}. 

Training such models requires correspondingly large machines. In 2012, the breakthrough AlexNet paper~\cite{krizhevsky2012imagenet} trained with model parallelism over two GPUs. That same year, Google harnessed their datacenter-scale CPU clusters to train asynchronously in the DistBelief system~\cite{dean2012large}. The Deep Learning revolution sparked huge investments in GPUs: NVIDIA revenues rose an average of 50\% year-over-year every quarter from mid-2016 to mid-2018~\cite{nvidiarevenue}. By 2015, Google had built a specialized neural network accelerator, the Tensor Processing Unit (TPU), a single chip which offered over a 10x improvement in performance/watt, peak performance, and inference latency~\cite{jouppi2017datacenter}. Within two years, Google’s second-generation TPU used 256-chip pods to train a single model with near-perfect parallel scaling~\cite{jouppi2020domain}; the third-generation TPU increased pod size to 1024~\cite{jouppi2020domain,kumar2019scale}. NVIDIA and other GPU suppliers have fielded clusters of similar scale, with Microsoft and OpenAI constructing a 10,000-GPU cluster~\cite{microsoft2020}. The space-race uses increasingly accurate models to approach Artificial General Intelligence, but there is no doubt that the hardware being fielded is also astronomically ambitious. 

\begin{figure}[t]
\begin{center}
\includegraphics[width=1\columnwidth]{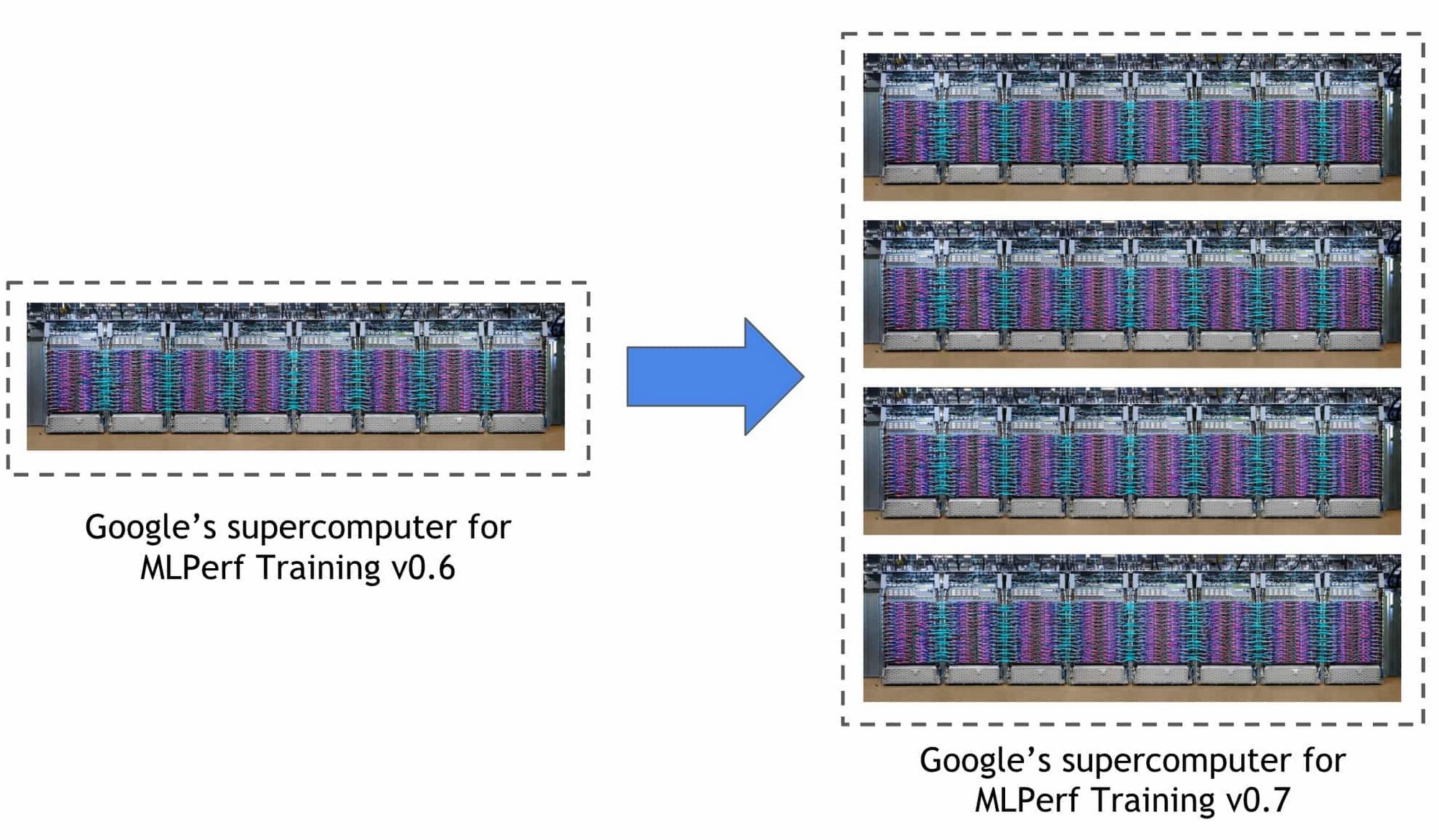}
\caption{TPU-v3 1-pod vs 4-pods in the Google datacenter.}
\vspace{-1em}
\label{fig:multipod-scaleout}
\end{center}
\end{figure}

Unlike the space race, where low-earth orbit and the moon make for obvious milestones, the best way to measure the accomplishments of these parallel machines is less concrete. Benchmarking competitions can serve this purpose: AlexNet surprised and transformed the vision community by winning the ImageNet Large-Scale Visual Recognition Competition~\cite{russakovsky2015imagenet} in 2012. Computer architects and system builders recognized the need for a benchmark suite similar to SPEC and TPC in their field, and a broad coalition of universities and companies founded MLPerf in 2018 to serve this need~\cite{mlperf}. In particular, the MLPerf Training division~\cite{mattson2019mlperf} attracts HPC-scale entries, as submissions compete to reach state-of-the-art accuracy on parallel training problems on massively parallel clusters in minimum wall-clock time. The techniques used in MLPerf submissions generally benefit the deep learning community, as they are folded into systems, libraries, compilers, and best-practices application code. This paper focuses on Google’s MLPerf 0.7 Training submission, and explains the algorithmic, architectural, performance, and system-tuning techniques that demonstrated world-class training at scale.

MLPerf~\cite{mattson2019mlperf} is a machine learning benchmark suite that is designed to benchmark different classes of ML accelerators and frameworks on state-of-the-art ML tasks. It has gained industry wide support and recognition. The recently concluded MLPerf-v0.7 Training submission round has submissions from NVIDIA, Google, AliBaba, Fujitsu, Shenzhen Institute and Intel.  Along with CPUs and NVIDIA GPUs, benchmarked hardware included the Google TPU-v3 and TPU-v4 as well as an AI accelerator from Huawei. ML frameworks included PyTorch, TensorFlow, JAX, MXNet, MindSpore and Merlin HugeCTR.

Like systems benchmark suites which have come before it, the MLPerf benchmark suite is pushing performance forward and our MLPerf-v0.7 Training submission on Google TPU-v3 and TPU-v4 systems showcase the large scale we are able to achieve. The MLPerf-v0.7 rules add new models, namely: i. BERT, a large language model, ii. DLRM, a deep learning recommendation system, and iii. an enhanced larger version of MiniGo to achieve higher scalability. An MLPerf training benchmark involves training a model (e.g., BERT) on a specific dataset (a Wikipedia dump) to a pre-defined convergence test metric while following specific methodology for parameters, optimizations, and timing.

\begin{figure}[t]
\begin{center}
\includegraphics[width=1\columnwidth]{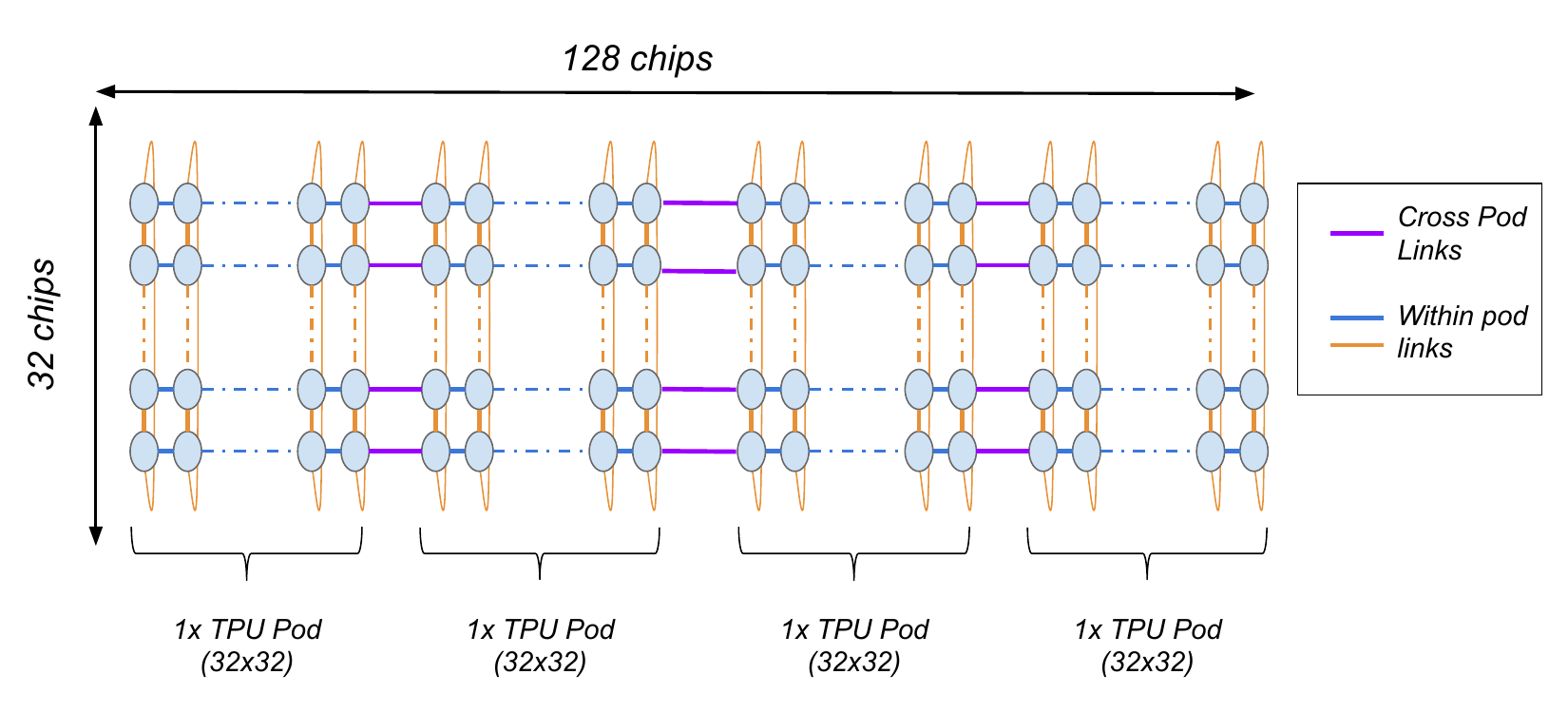}
\caption{TPU-v3 4-pod configuration where cross-pod links connect neighboring TPU-v3 pods in the Google datacenter.}
\vspace{-1em}
\label{fig:multipod-network}
\end{center}
\end{figure}

In order to explore the limits of concurrency in the MLPerf models we assembled a TPU-v3 Multipod with 4096 chips, with 105 TFLOPS per chip at peak. It is four times larger than the TPU-v3 pod used for the MLPerf-v0.6 training benchmark submission.  A 4-pod Multipod configuration with 4096 TPU-v3 chips is shown in~\Fig{fig:multipod-scaleout}. Here the two pods are connected along the X-dimension of the mesh by the cross-pod optical links (~\Fig{fig:multipod-network}). These links are longer than standard TPU-v3 within-pod links. The MLPerf benchmarking was done on a 4-pod Multipod with 4096 chips in a 128x32 2-D mesh topology (with within-pod torus links at the Y edges).  As the TPU-v3 chip had only 1024 entries in the routing table, we used a sparse routing scheme where only neighbors along rows and columns were visible to each chip. This was sufficient for achieving peak throughput in the all-reduce communication operations.

We chose a subset of MLPerf models to benchmark at the Multipod scale.  These included i) BERT, ii) ResNet-50, iii) Transformer and iv) Single Shot Detector (SSD). In BERT and ResNet-50 we used batch parallelism to scale to the Multipod, while in Transformer and SSD we used a combination of model parallelism and batch parallelism techniques.  The Mask-RCNN and DLRM models are also discussed in this paper. For the Mask-RCNN model, the available batch parallelism is extremely limited and we therefore present results on a slice with 512 TPU-v3 chips. For the DLRM model, scalability is capped by limited global batch size and communication overheads quickly outweigh scale-out benefits. We present results on a slice with 256 TPU-v3 chips.

This paper made the following major contributions.
\begin{itemize}
\item A world-record scale ML architecture with 4096 nodes. It was the biggest machine for MLPerf-v0.7 and it put extra pressure on the dedicated interconnect. This machine extends the X-dimension with cross-pod optical links that have higher latency and lower bandwidth than the links within pods. To mitigate the link speed difference, we designed a novel all-reduce algorithm that pushes most of the all-reduce payload along the Y dimension that results in high throughput as communication along X-dimension is reduced by a factor that is the same as the Y-dimension size (32).
\item Optimized global summation for model parallelism.
The current state-of-the-art MeshTF~\cite{shazeer2018mesh} maps language models along batch and model dimensions that are then mapped to the physical 2-D mesh of TPU-v3. We found this approach had significantly high communication overheads as the gradient all-reduce step is executed on a 1-D ring. We present a novel strided communication optimization scheme that enables high throughput in both the forward and the gradient reduction steps, that results in the MLPerf Transformer model training in 16 seconds.
\item We scale weight update sharding (distributed optimizer) in a complex hybrid of data and model parallelism scenario via model parallelism and spatial partitioning.
\item Analysis of the JAX programming model and comparison with TensorFlow. This is the first paper that studies JAX at scale, uses JAX on TPU (multi)pods, and uses model parallelism techniques (SPMD partitioning and weight update sharding) in JAX. The JAX results demonstrate the generality of TPUs and the enhancements added to XLA, and provide a useful comparison for multi- vs. single-controller design points in distributed ML systems.
\item Multipod Performance Results. Four models finish training in under 30 seconds. BERT and DLRM, the models recently added to MLPerf-v0.7, are optimized at a TPU Multipod scale for the first time.
\end{itemize}

%% file: frameworks.tex
\section{Multiple Frameworks}
\label{sec:frameworks}

While the primary frontend for TPUs has historically been TensorFlow~\cite{abadi2016tensorflow}, the hardware and XLA compiler are general enough to support other programming environments. Therefore in this paper, we chose to benchmark both TensorFlow and JAX~\cite{frostig2018compiling}, a new, research-oriented numerical computing system based on XLA~\cite{xla}. Both systems required additional software engineering to scale effectively to the Multipod, but they ultimately achieved similar benchmark results.

\begin{figure}[t]
\begin{center}
\includegraphics[width=1\columnwidth]{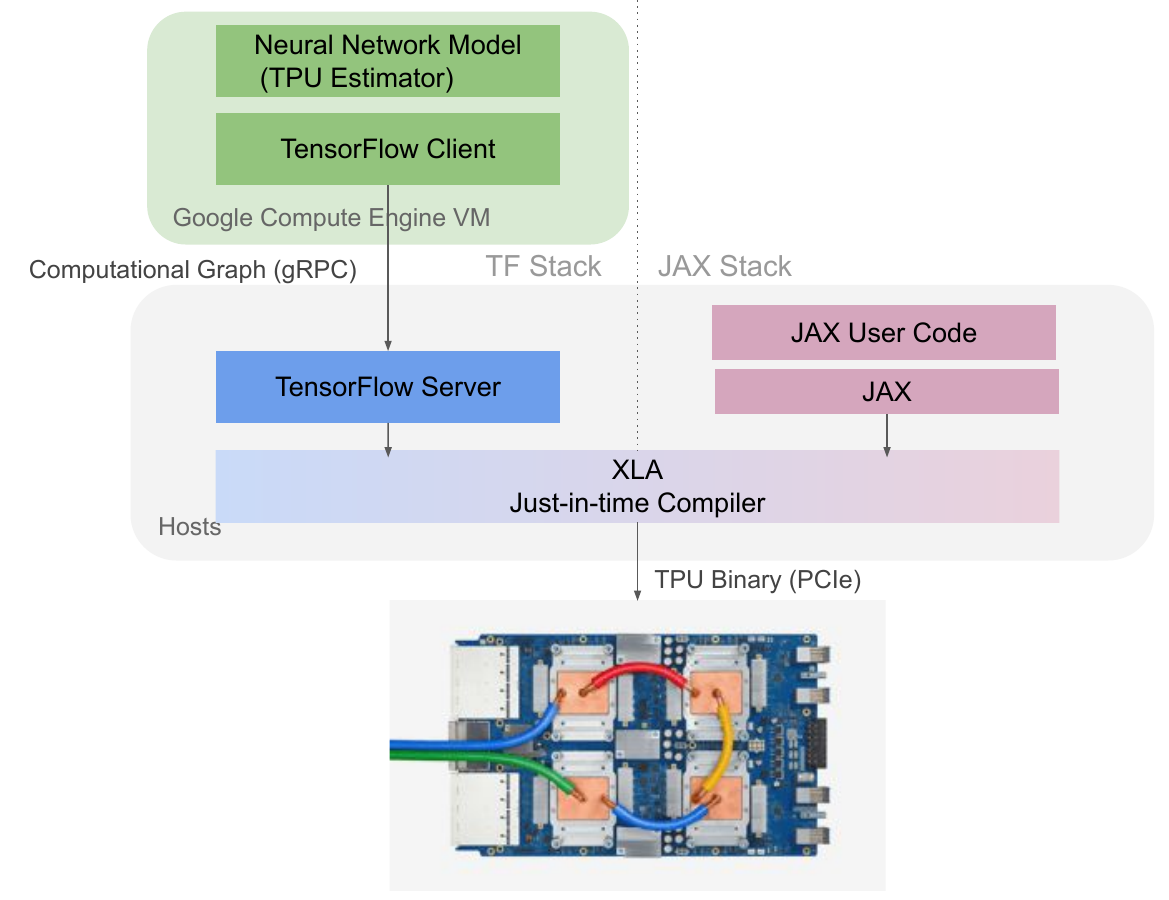}
\caption{Stack view of the TF and JAX frameworks on the TPU-v3 machines.  }
\vspace{-1em}
\label{fig:tpu_stack}
\end{center}
\end{figure}

As shown in~\Fig{fig:tpu_stack}, two architectural differences between TensorFlow and JAX differentiate their performance at scale. First, they have different staging approaches. TensorFlow embeds an expressive and dynamic intermediate language (TensorFlow graphs that can span both accelerators and CPU hosts) in Python, and then JIT-compiles subsets of these graphs with XLA. Meanwhile, JAX has one fewer stage: it is a staged programming environment that embeds JIT-compiled XLA programs (for static compiled performance on accelerators and parallelism on the accelerator network) in the Python host language (used for dynamic and on-accelerated computations). As a consequence, TensorFlow has additional compilation steps, which we accelerated using multithreading, while JAX requires more careful management of Python bottlenecks (for instance, moving blocking tasks like data infeed off of the main thread).

Second, they enable different distributed programming models. JAX adopts a multi-client approach to distributed programming, running a separate copy of the same JAX code (including the Python interpreter) on each host in the pod. The programs communicate with each other in only two ways: at startup time, to coordinate TPU mesh setup, and in XLA-compiled collectives such as all-reduce that operate over the dedicated TPU network during model training. On the other hand, TensorFlow programs TPUs with a single-client approach, giving one Python process (running either on one of the hosts in the pod or elsewhere) global visibility and control over the entire distributed system. The rest of the TPU hosts run a TensorFlow server that executes partitioned subsets of TensorFlow graphs sent via RPCs from the client over the datacenter network.

These two approaches differ in usability and performance characteristics. While TensorFlow’s single-client distributed system enables user code that directly reflects the overall workload, JAX’s multi-client approach enables more direct control of the code that runs on each worker. JAX invokes the XLA compiler independently on each host—relying on deterministic compilation to avoid incompatibilities between the resulting programs—while TensorFlow compiles once and distributes the binaries to the workers. The TensorFlow representation of multi-device graphs can also cause Amdahl’s law bottlenecks, as the client process incurs graph construction and optimization time proportional to the number of workers, while JAX setup times (other than TPU topological mesh initialization) do not change significantly with an increase in the number of workers.

%% file: techniques.tex
\section{Scalability Techniques}
\label{sec:scalability}

In this section, we describe the optimization techniques required to scale MLPerf-v0.7 models implemented in both frameworks to the 4096-chip TPU-v3 Multipod machine. Optimization of MLPerf-v0.6 models to a single TPU-v3 pod is presented in~\cite{kumar2019scale}. To achieve higher scale on the Multipod, we next present novel all-reduce optimizations, aggressive model parallelism and input pipeline optimizations. 

\subsection{Model Parallelism}

In models where data parallelism is limited, we use model parallelism to achieve higher concurrency on TPU-v3 Multipod.  We leverage XLA’s Single Program Multiple Data (SPMD) partitioner~\cite{lepikhin2020gshard} to automatically partition model graphs based on light-weight annotations.  In the segmentation models, SSD and MaskRCNN, we implement spatial partitioning by annotating input images. The SPMD partitioner can automatically parallelize computation along the spatial dimensions.  These models have relatively large spatial dimensions  (8000x1333 for MaskRCNN and 300x300 for SSD). The SPMD partitioner inserts halo exchange communication operations to compute the activations for the next step from spatially partitioned computations. Both of these models enable spatial partitioning along 8 cores to achieve the highest level of concurrency. Communication optimization and elimination of Amdahl bottlenecks via the XLA compiler SPMD approach~\cite{lepikhin2020gshard} enabled higher concurrency in spatial partitioning. For example, in MaskRCNN the largest batch size is 256, but we were able to parallelize the training on up to 1024 accelerator cores. 

In the language models such as the MLPerf transformer benchmark, where the spatial dimensions are small, we explore partitioning the feature dimension as described in~\cite{shazeer2018mesh}, but implemented as annotations for the SPMD partitioner.  In this approach, the model weights and activations are split on a tile of the TPU mesh. In the forward pass, partial matrix multiplication operations are computed on each core of the tiled sub-mesh.  The activation contributions from each core are reduced via an all-reduce operation on the tiled submesh to execute the next layer of the model.  The backward pass has a similar partial matrix multiplication followed by all-reduce producing both activations and gradients. As the weights are also partitioned, the gradients are summed between a partitioned core and its corresponding peer on every other tiled sub-mesh of the TPU machine. Techniques to optimize gradient summation on Multipod are presented in Section 3.3. 

\subsection{Weight Update Sharding}
In traditional data parallelism, model weights are replicated and updated by the optimizer at the end of each training step. However, this computation can become significant when the mini batch size per core is small. For example, we measured in the MLPerf BERT model, the LAMB optimizer weight-update time is about 18\% of the step time on 512 TPU-v3 chips. The weight-update-sharding technique~\cite{xu2020automatic} distributes this computation by first executing a global reduce-scatter after which each accelerator has a shard of summed gradients.  This is used to compute a shard of updated weights. In the next step, the shard of updated weights is globally broadcast to update all replicas. To achieve higher speedups we enable weight update sharding in both data and model parallelism. In the segmentation models, where the weights are replicated, the weight-update-sharing scheme is similar to data parallelism.  However, when the weights are distributed, we execute multiple concurrent weight-update-sharding steps in each model parallel core across all the replicas. 

\subsection{Optimized Global Summation}

The gradient summation step is critical to achieve strong scaling with MLPerf benchmarks~\cite{mattson2019mlperf}.  In order to optimize gradient summation on the large TPU-v3 Multipod, we take advantage of the torus wrap links along the Y-dimension. A bidirectional ring is used to execute a reduce-scatter operation along the Y-dimension with the output being a shard of the summed gradients along the Y-ring. Next, a reduce-scatter is executed along the X-dimension. This is followed by a weight update computation with the gradient shard as the input. The updated weights are broadcast first along X and then Y in two steps.  Note, in data parallelism, the payload transferred along the X-dimension is 32 times less than the data transferred along the Y-dimension. 

\begin{figure}[t]
\begin{center}
\includegraphics[width=1\columnwidth]{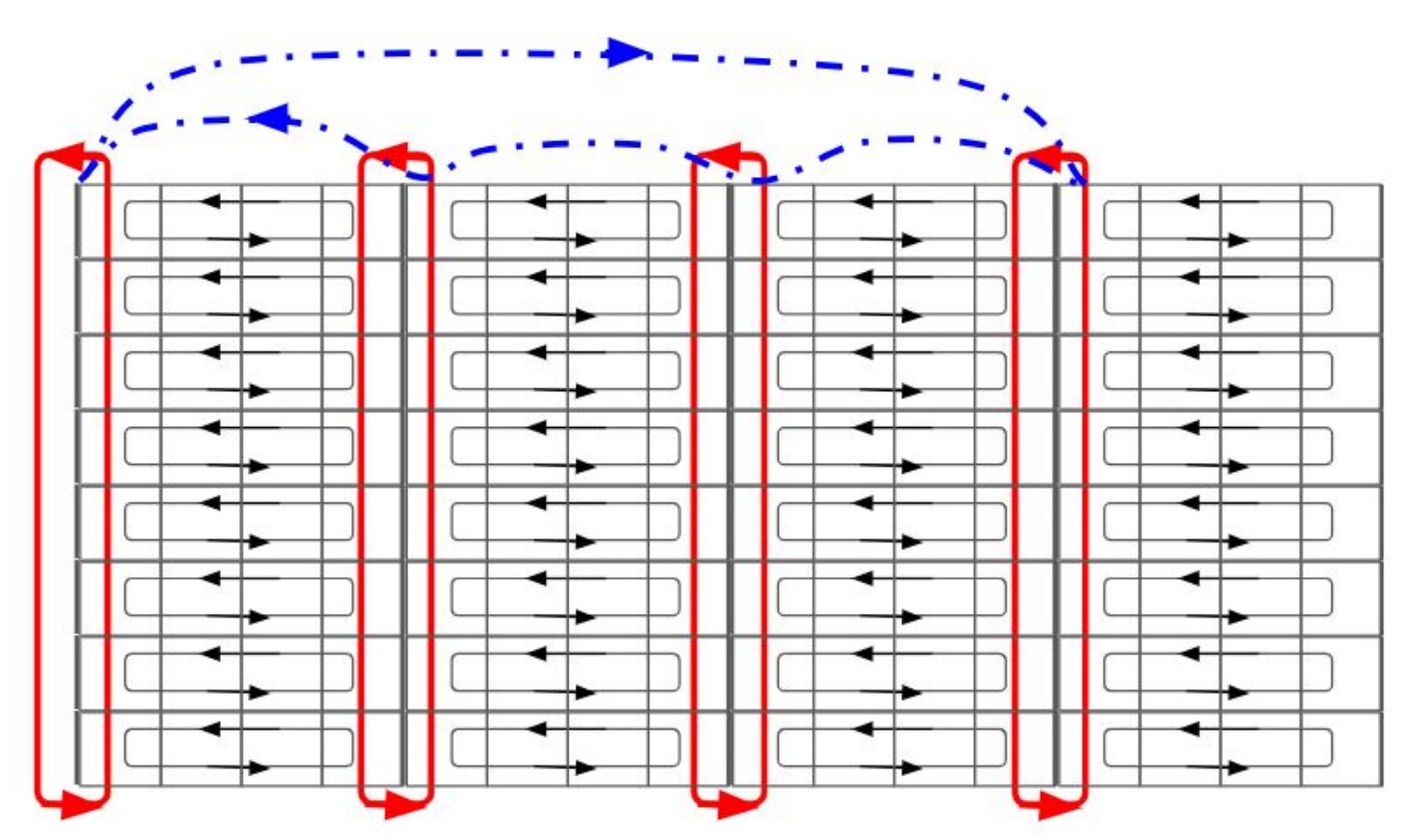}
\caption{Figure shows a 16(mesh) $\times$ 8 (torus) with model parallelism along 4 chips. Three different ring reductions are shown here.  i) a black ring reduction for the model parallel forward pass. ii) red rings do the bulk of the gradient reduce scatter along the Y dimension iii) the dotted blue line shows gradient reduction among model peers (only peer id = 0 is shown).
}
\vspace{-1em}
\label{fig:global_summation}
\end{center}
\end{figure}

In the MLPerf transformer benchmark, we execute distributed matrix multiplication operations by sharding model weights on up to 4 neighboring TPU cores. These cores are placed along a line on the X-dimension. In the forward pass of ML training, all-reduce calls are executed along short rings of X-neighbors. The gradient summation on the Y-dimension stays unchanged as with data parallelism. However, the gradient summation along the X-dimension hops over peers that are model parallelism neighbors.  The different ring reductions in the Transformer benchmark are illustrated in~\Fig{fig:global_summation}. In the BERT and transformer models, we also used the brain-float 16-bit floating point precision (bfloat16) to further reduce gradient summation overheads.

\subsection{Distributed Computation of Evaluation Metrics}

The train and evaluation computations are executed in a tight loop on the TPU accelerators.  The result of the train loop updates model weights in the HBM storage on each TPU accelerator.  The updated weights are then used to evaluate the output metrics for the number of epochs specified in the MLPerf rules.  In benchmarks where the evaluation batch size is larger than the number of examples in the evaluation dataset, the evaluation dataset is padded with dummy examples. In the TensorFlow implementation, the eval output tensors are used to compute the evaluation metric (for example top-1 accuracy in the Resnet-50 benchmark) on the TPU master host. However, in the JAX implementation the computation of the evaluation quality metric is fully distributed via global summation calls. 

\subsection{Input pipeline optimizations}
\label{sec:scalability:input}
One of the challenges of scaling the ResNet-50 model is load-imbalance in the host input pipeline.  With the massive scale of a Multipod, some host input pipelines have high overheads of decompressing large JPEG images. Our solution is to store uncompressed images in the host memory so that the host input pipelines execute only i) random crop, ii) random flip and iii) image normalization with a constant mean and variance as specified in the MLPerf reference. This significantly increases the throughput of the host input pipeline allowing it to create a large prefetch buffer. So, when the host pipeline preprocesses a large input image it can feed TPUs with images in the prefetch buffer, thus, eliminating the input pipeline load-imbalance on the Multipod system. This optimization increases the training throughput of ResNet-50 by 35\% on a Multipod. 
With uncompressed images, although the need for memory capacity increases, the available memory capacity in the system is sufficient. The Multipod has about a thousand CPU host servers and the input is sharded across all these. Using uncompressed images does not incur extra memory throughput overhead, since decompressing images in host memory results in more memory transfers.

For BERT, one of the key techniques to improve convergence is to guarantee randomness and coverage in data shuffling. We find two things very helpful for BERT, using the tf.data.shuffle function before the tf.data.repeat function at file level, and increasing the shuffle buffer size at sequence level. At file level, proper data shuffling is especially important as the system scale increases, where every host has fewer data files to work with. For example, the 500 files in the BERT reference model will result in a medium-scale system with 128 hosts having only about 4 files per host. Executing a tf.data.repeat before tf.data.shuffle gives better randomness and coverage of the whole dataset, where the former guarantees the stochasticity, and the latter guarantees the model catches all information available in the dataset. At sequence level, shuffling with small buffer size incurs large run-to-run convergence difference, which originates from the difference of biased training batch at each training iteration, leading to very different convergence trajectories of different runs. With larger buffer sizes, every training batch of different runs can be more uniformly sampled from the whole dataset, which therefore reduces run-to-run difference. 

DLRM, like many other recommendation engines, can quickly become input bound as the model accommodates a large per-core batch size while having a small step latency.  One key input pipeline optimization for such models is to use host parallel processing to parse data at batch granularity, instead of per-sample.  In the case of the dataset used for this model, each training sample is composed of about 40 input features.  An additional optimization is to transmit input features over the PCIe bus in a stacked form, reducing the overhead of transmitting many features separately.  Finally, batching overhead can be mitigated by shuffling and pre-serializing data in batch form.

%% file: model.tex
\section{Model Optimizations}
In addition to the optimizations mentioned previously, in this section we present the optimizations applied to each MLPerf model.  With the exception of MaskRCNN and DLRM, all other models are implemented in both TF and JAX.  Note, the JAX implementations use the same scalability and convergence techniques as TF models, resulting in very similar step times as well as number of convergence steps. There are subtle differences w.r.t. TF implementations due to JAX’s multi-client design. For example, JAX allows initializing datasets and input pipelines concurrently in each TPU worker. Also, the global evaluation accuracy is computed via a global all reduce operation on TPUs, compared to TF, where the coordinator CPU process computes the sum after gathering all local accuracy metrics via host RPC calls. 

\subsection{BERT}
BER~\cite{devlin2018bert} with the wikipedia dataset is newly added in MLPerf-v0.7. It is a pre-training task for language understanding with bi-directional transformer architecture. Thanks to the LAMB optimizer~\cite{you2019large}, BERT can scale very well to large batch sizes, and we are able to use data parallelism at a 4096-chip scale. Scaling BERT in large systems involves optimizations of two aspects, step time and steps to converge. 

Other than the optimizations in Section 3, at model level, we optimize the step time of BERT by reducing the stress on architectural bottlenecks, including memory bandwidth, vector units, and registers allowing computation to be executed on the TPU-v3 matrix units with minimal pipeline bottlenecks. To reduce memory bandwidth, we utilize the bfloat16 data type~\cite{bfloat16} for model activations and gradients aggregation, which largely improves the step time and does not have negative effects on model convergence. To reduce the stress on vector units, we move the scalar multiplications and divisions to the smaller side of matrix multiplication by leveraging the commutativity of scalar multiplication and matrix multiplication. To reduce register spilling, we combine small variables, such as layernorm variables, into one large TensorFlow tensor. This largely reduces the number of variable addresses to store in the registers and therefore speeds up the training step time.

To reduce the steps to converge, we optimize hyperparameters and data shuffling in the input pipeline. First, we use Google Vizier~\cite{golovin2017google} to fine tune the hyperparameters for large batch training, enabled by the scalability of LAMB optimizer. This allows us to leverage maximum data parallelism, which gives better time to accuracy compared to model parallelism. Second, as detailed in Section~\ref{sec:scalability:input}, we optimize the way data is shuffled in the data input pipeline in order to ensure the convergence in large systems. This is critical for guaranteeing the stochasticity of the optimizer because large systems with thousands of hosts typically assign less data to each host.

\subsection{ResNet-50}

ResNet-50~\cite{he2016deep} is one of the most widely-used models for ML benchmarking. MLPerf uses the ResNet-50 model with the ImageNet-1K~\cite{russakovsky2015imagenet} dataset as the image classification benchmark. Specifically, MLPerf uses the variant termed ``version 1.5''~\cite{goyal2017accurate} to indicate a slight modification to the original model architecture which is commonly found in practice. In order to scale the ResNet-50 benchmark to the TPU-v3 Multipod system, we apply optimizations including distributed evaluation, distributed batch normalization, weight update sharding, and gradient summation. The MLPerf-v0.7 reference model uses the LARS optimizer~\cite{you2017large} that adaptively scales learning rates, which enables training ResNet-50 with data parallelism on large batch sizes.  After the momentum hyperparameters are tuned, we are able to finish training in 88 epochs with batch 65536 on the Multipod. 

\subsection{Transformer}

Transformer represents the state-of-the-art language translation in the MLPerf suite and is one of the two translation models. Trained on the WMT English to German dataset, Transformer uses an attention-based model which differentiates it from the other language model in MLPerf, GNMT. It has been observed that it is hard to scale Transformer with a fixed epoch budget beyond a global batch size threshold given the current dataset~\cite{shallue2018measuring}. Therefore both data and model parallelism are applied to scale the Transformer model to a TPU-v3 Multipod system. With model parallelism, the model is able to run with fewer than batch one per core, using a fixed global batch size of 2048 where the hyperparameters have been well tuned.

SPMD sharding is employed to enable model parallelism. Unlike spatial partitioning (sharding the images) nor Gshard~\cite{lepikhin2020gshard} (which has sparse components and all-to-all communications), dense sharding is applied to the Transformer model. Shared embedding layers, multi-heads attention projection layers and feed-forward layers are sharded, along with vocab, num\_heads, and hidden dimensions, respectively. To speed up gradient all-reduce, 2D cross-replica all-reduce is enabled for SPMD sharding with X-dimension hops over model parallelism neighbor replica. The all-reduce communication is performed in bfloat16 floating point precision to further improve the performance.

\subsection{SSD} 
Single Shot Detection (SSD) is one of two image segmentation models in MLPerf; SSD is intended to reflect a simpler and lower latency model for interactive use cases such as in end-point and non-server situations. Notably, SSD uses a pre-trained ResNet-34 backbone as part of the architecture. The MLPerf SSD benchmark is trained on the COCO dataset~\cite{lin2014microsoft}. In the MLPerf-v0.6 submission we had used a global batch size of 2048 and 4-way model parallelism. In this round of MLPerf submissions, we are able to train with a batch size of 4096 using new hyperparameters. Note, this is still much smaller than the batch size of 65536 available in the ResNet-50 model. We used XLA’s SPMD partitioner to enable scaling up to eight TPU cores via model parallelism, replacing XLA’s MPMD spatial partitioner used in MLPerf-v0.6.  SPMD has better scalability in compilation time and enabled us to increase the largest scale for SSD training from 2048 TPU-v3 cores in MLPerf-v0.6 to 8192 cores in MLPerf-v0.7.   A unique benefit of the SPMD partitioner is that it enables the weight-update-sharding optimization even with model parallelism, that results in a 10\% speedup.  It is challenging to get high speedups in the SSD model from model parallelism as there are communication overheads from halo exchange and load imbalance as different workers may get uneven tiles of work. In addition,  the input image to the SSD model is relatively small (300$\times$300) in the first layer and is further reduced to 1$\times$1 in the last layer.  In spite of the above, we are able to get speedups on up to 8 TPU cores used for spatial partitioning.

The SSD implementation with JAX also follows the similar scalability and convergence techniques as its TF counterpart. In addition to the listed differences,  the COCO eval execution is slightly different. In TF SSD, the results of the predictions are all brought to the TF coordinator process via host calls, and COCO eval is executed by the TF coordinator process’s CPUs. Since JAX does not have a separate coordinator process, COCO eval is executed on the worker processes in a round robin fashion to improve the load-imbalance, e.g., first worker executes the first COCO eval, the second worker executes the second one, and so on.

\begin{table*}[h]
  \centering
  \begin{tabular}{lcccc}
    \toprule                \\
    Benchmark & TPU-v3 Chips & TF Runtime (mins.) & Speedup over MLPerf-v0.6 & JAX Runtime (mins.) \\
    \midrule
    Resnet-50 & 4096 & 0.48 & 2.67 & 0.47 \\ 
    BERT & 4096 & 0.39 & N/A & 0.4 \\
    SSD & 4096 & 0.46 & 2.63 & N/A \\
    SSD & 2048 & 0.623 & 1.94 & 0.55 \\
    Transformer & 4096 & 0.32 & 2.65 & 0.26 \\
    MaskRCNN & 512 & 8.1 & 4.4 & N/A \\
    DLRM & 256 & 2.4 & N/A & N/A \\
    \bottomrule
\end{tabular}
\caption{End-to-end time achieved with TF 1.x and JAX MLPerf-v0.7 benchmarks on the TPU Multipod machine. The table also presents the speedups achieved over the MLPerf-v0.6 TF submissions.}
\label{tab:mlperf_v_07}
\end{table*}

\begin{table}[h]
  \centering
  \begin{tabular}{lcc}
    \toprule                
Benchmark & TPU-v3 TensorFlow & TPU-v3 JAX \\
          & (TPU Chips)   & (TPU Chips) \\
    \midrule 
Resnet-50 & 8.30 (4096) & 2.23 (4096) \\ 
BERT & 17.33 (4096) &  3.17 (4096)  \\
SSD & 12.87 (4096) & 2.03 (2048) \\
Transformer & 14.47 (4096) & 4.90 (4096) \\
    \bottomrule
\end{tabular}
\caption{Initialization time (minutes) comparison between TensorFlow and JAX. Because of the multi-client distributed system of JAX, it shows lower initialization time than TensorFlow, which compiles a multi-device graph on the master host.}
\label{tab:mlperf_v0_7_init_times}
\end{table}

\subsection{MaskRCNN}

Mask-RCNN is the heavy weight object detection benchmark in MLPerf. Besides object detection, Mask-RCNN also performs instance segmentation, which assigns a semantic label as well as an instance index to each pixel in the image. Unlike SSD, which is a one stage detector, Mask-RCNN has two stages: one for proposing instance candidates and the other for fine-tuning the proposals. Also, Mask-RCNN uses a larger image size (800$\times$1333) over SSD (300$\times$300) even though they both train the COCO dataset. Furthermore, Mask-RCNN uses a Resnet-50 backbone plus Feature Pyramid Network contrasted with SSD's use of Resnet-34. 

Scaling MaskRCNN training is challenging as the largest batch size that achieves the MLPerf model quality is quite small.  In MLPerf-v0.6 it was trained with a batch size of 128 while in MLPerf-v0.7 we are able to increase the batch size to 256 with new hyperparameters.  Data parallelism is used uptill 128 TPU cores and then model parallelism is applied to scale further 1024 TPU cores. The following optimizations are enabled in XLA’s SPMD partitioner to achieve high throughput on TPUs:
\begin{itemize}
\item Model parallelism optimized Gather: ROIAlign operation in Mask-RCNN's is dominated by non contiguous gather operations. These are optimized by one-hot-matmul calls that execute on the TPU matrix unit achieving  linear speedups when increasing the number of model parallelism partitions.
\vspace{-1em}
\item Resharding: the convolutions are split on spatial dimensions. However, when executing the einsum computation, the inputs are re-partitioned on the non-contracting dimension with minimal communication overheads
\vspace{-1em}
\item Partitioning more ops: there was no XLA partitioner support for some mask-RCNN operations such as top-k, gather and special case convolutions, which can become an Amdahl bottleneck. We  added support in the XLA compiler to aggressively split the computation between the TPU cores and increase speedup.
\vspace{-1em}
\item Communication optimization: there are significant communication overheads between the model parallelism peer cores from resharding, gradient reductions and halo exchange in convolutions. We optimized these in the XLA compiler to reduce communication overheads from 30\% to about 10\%. Optimizations include minimizing the number of resharding steps, executing a single gradient all-reduce across model cores and replicas and barrier optimizations for the halo exchange.
\end{itemize}

\subsection{DLRM}
The Deep Learning Recommendation Model (DLRM)~\cite{naumov2019deep} is a neural-net based recommendation engine that is newly added in MLPerf-v0.7.  The task of the recommender is pCTR click prediction using a combination of categorical and integer features and a single binary label.  The MLPerf dataset is the open-source Criteo Terabyte click logs dataset~\cite{criteo}, a corpus of over 4B examples.  Embedding table lookups and updates play a significant role in performance due to large unique vocabulary sizes for some of the categorical features.  The remainder of the model is dominated by fully connected layers.

We use a global batch of 65536, the maximum with converging hyperparameters, in order to enable scalability. Despite the larger batch size, scalability is still limited for this problem, as step latency is small and communication overheads become a significant portion of runtime.  As a result, we do not use a full multi-pod for this model, but rather a fraction of a pod.  The input pipeline is also challenging due to the large batch size and small step latency.  Section 3.5 discussed relevant input optimizations. Other optimizations used for DLRM include:
\begin{itemize}
\vspace{-1em}
\item Partition large embedding tables:  This is actually necessary to run the model, due to the large memory footprint of the tables.  The optimization involves choosing to replicate small tables and partition large ones.
\item Optimize gather overheads:  The model includes a feature self-interaction function that uses a gather to eliminate redundant features.  We mask the redundant features with zeros and modify the downstream fully connected layers to ignore the null features during initialization.
\vspace{-1em}
\item Evaluate multiple steps without host communication:  The inference step latency of the model is small enough that the PCIe communication with the host and network gather poses an unacceptable overhead.  Instead, we perform multiple inference steps on device and accumulate them.
\vspace{-1em}
\item Custom evaluation metric library:  The evaluation metric is AUC (ROC) on a dataset composed of ~90M samples.  Popular python libraries scale poorly to this size, requiring ~60 seconds per metric computation on a modern workstation.  We write a custom C++ CLIF-wrapped implementation that relies on multithreaded sorting and loop fusion to compute the metric in ~2 seconds per call.
\end{itemize}

%% file: performance.tex
\section{Performance Analysis}

\begin{figure}[t]
\begin{center}
\includegraphics[width=1\columnwidth]{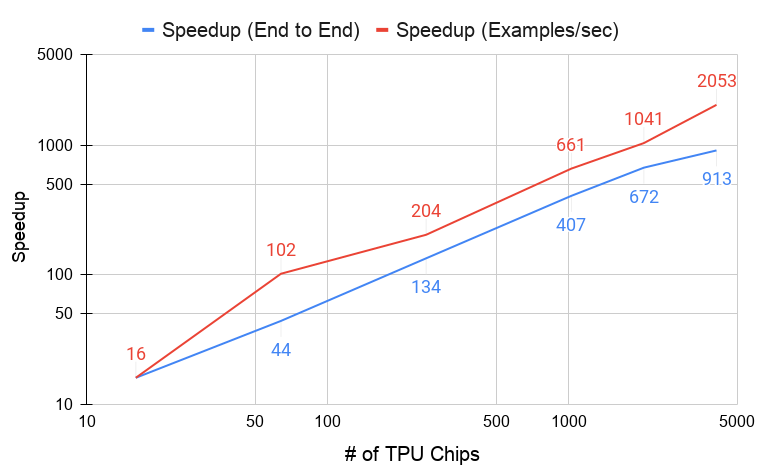}
\caption{Speedup of ResNet-50 with the number of TPU chips.}
\vspace{-1em}
\label{fig:resnet50_speedup}
\end{center}
\end{figure}

\begin{figure}[t]
\begin{center}
\includegraphics[width=1\columnwidth]{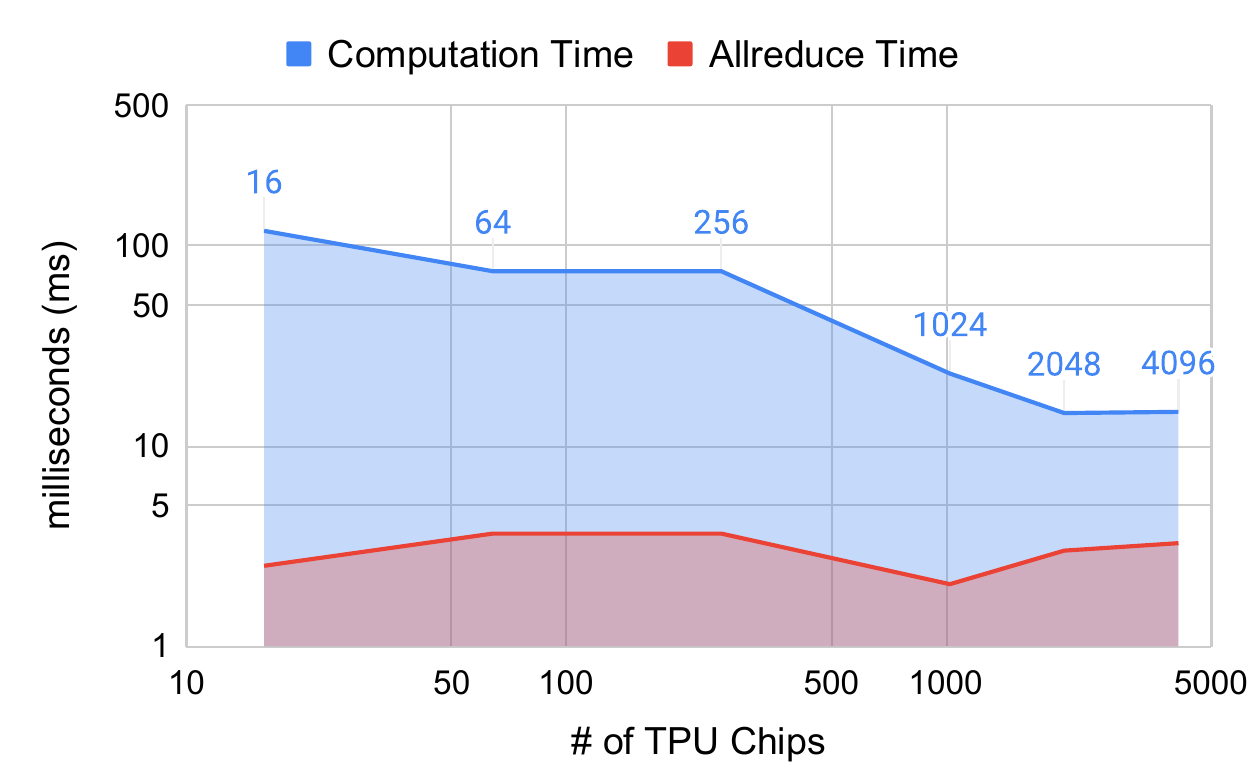}
\caption{TF ResNet-50’s computation and communication (all-reduce) time on TPUs for executing a single min-batch. Note, the mini-batch size is decreased from 256 to 16 per TPU chip as the scale is increased.}
\label{fig:resnet50_breakdown}
\end{center}
\end{figure}

\begin{figure}[t]
\begin{center}
\includegraphics[width=1\columnwidth]{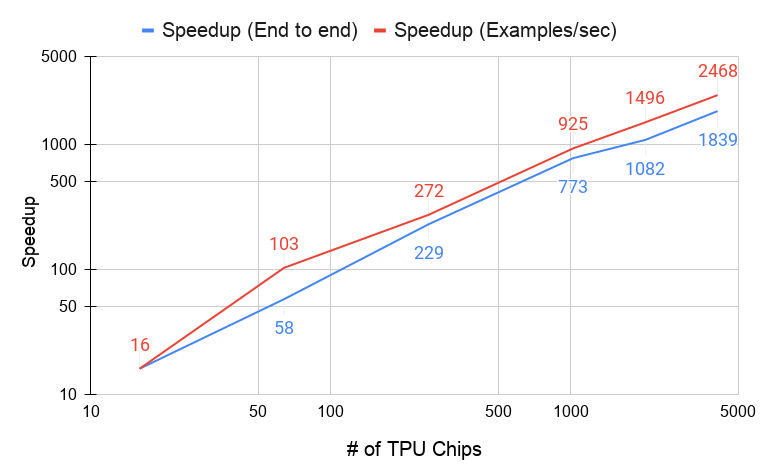}
\caption{Speedup of TF BERT with the number of TPU chips.}
\vspace{-1em}
\label{fig:bert_speedup}
\end{center}
\end{figure}

\begin{figure}[t]
\begin{center}
\includegraphics[width=1\columnwidth]{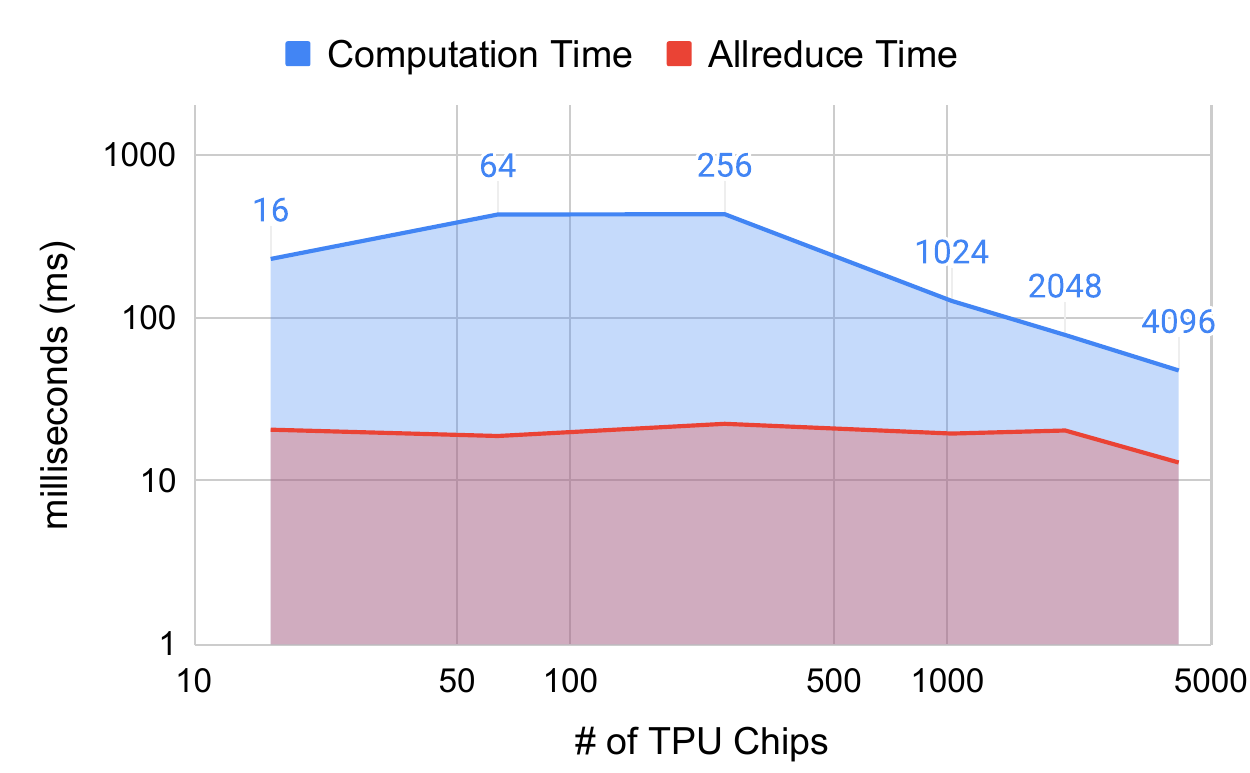}
\caption{TF BERT’s computation and communication (all-reduce) time on TPUs. Note, mini-batch per TPU chip is 2 at the 4096 chip scale and varies between 4 and 48 at other chip counts.}
\label{fig:bert_breakdown}
\end{center}
\end{figure}

In this section, we show the performance we are able to achieve with all the framework, infrastructure and model optimizations discussed.  Table~\ref{tab:mlperf_v_07} presents the end-to-end times in six MLPerf-v0.7 benchmarks on Google TPU-v3 Multipod. End-to-end time is defined as the time from data touch to computation of the evaluation metric that reaches target quality. Note that the presented optimizations and the larger number of accelerators in the Multipod result in four benchmarks training in under half a minute. The table also presents speedups over the Google TPU submission to the MLPerf-v0.6 benchmarks. Note that for DLRM, the best result of 1.21 minutes was achieved on a TPU-v4 machine.  As TPU-v4 is not the focus of this paper and it is not discussed here, the TPU-v3 result (2.4 minutes) is presented in its place.

Table~\ref{tab:mlperf_v0_7_init_times} compares the initialization time of TensorFlow and JAX on large-scale systems with 4096 and 2048 TPU chips. TensorFlow’s initialization time ranges from 498 seconds to 1040 seconds, while that of JAX’ is much lower, ranging from 122 seconds to 294 seconds. This is because of their different distributed programming models, which are tradeoffs between performance and usability, as mentioned in Section~\ref{sec:frameworks}. JAX uses a multi-client distributed system, where each client has one copy of the code and compiles its own graph. This generates constant initialization time for varied sizes of systems. On the other hand, TensorFlow gives one Python process global visibility and control of the whole system, which makes it easier for user code to be reflected on the whole workload. But this Python process needs to compile a large multi-device graph for the whole system, which takes longer for larger systems.

Our performance of BERT, ResNet-50, Transformer, MarkRCNN, DLRM and SSD outperform the competitors. \Fig{fig:resnet50_speedup} shows the end-to-end and throughput speedup of ResNet-50 with the number of TPU chips. It is not surprising that the throughput speedup is closer to ideal scaling than the end-to-end speedup. Note, at batch 64K the MLPerf ResNet-50 model takes 88 epochs to train, while at batch 4K it only needs 44 epochs to train to the target top-1 accuracy of 75.9\%.

To further examine the Amdahl’s Law bottleneck on large systems, \Fig{fig:resnet50_breakdown} breakdowns the computation and all-reduce (communication) overhead in the step time, where the blue area shows the computation time, red area shows the communication time, and the sum of the two is the step time on the device. Note that both x- and y-axes are in log scale. With scaling to larger systems, the computation time keeps decreasing while the communication time, i.e., the all-reduce time, stays almost constant. Using 4096 chips, the all-reduce operations take 22\% of the total device step time.

\Fig{fig:bert_speedup} shows the BERT speedup varying the number of TPU chips. BERT shows the highest scalability on systems with 16 to 4096 chips. \Fig{fig:bert_breakdown} shows the computation and communication time breakdown for BERT. Compared to ResNet-50, the Amdahl’s Law bottleneck for BERT takes larger percentages, in all scales ranging from 16 to 4096 chips. With the 4096-chip configuration, the all-reduce communication takes 27.3\% of the total device step time.

\begin{figure}[t]
\begin{center}
\includegraphics[width=1\columnwidth]{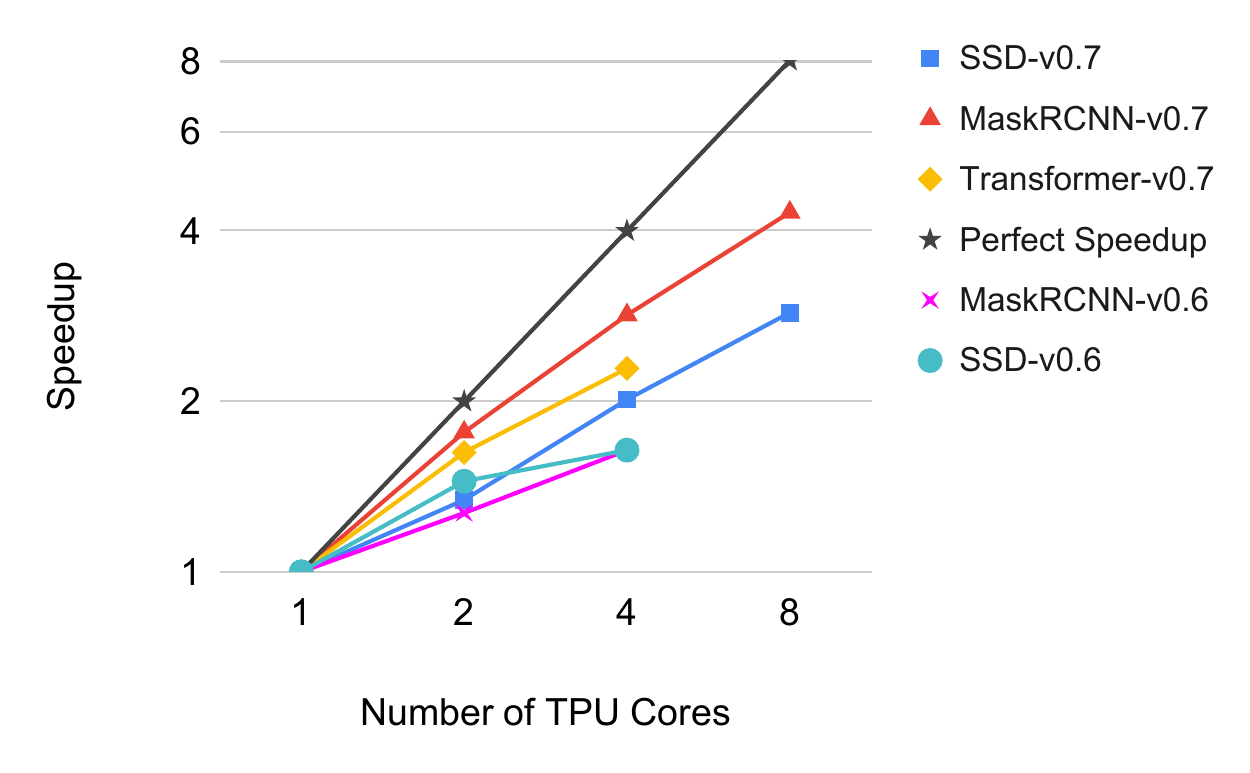}
\caption{Speedup via model parallelism in MLPerf-v0.7 TF. }
\label{fig:model_parallelism_speedup}
\end{center}
\end{figure}

\Fig{fig:model_parallelism_speedup} shows speedups via model parallelism in the SSD, MaskRCNN and Transformer benchmarks. Both SSD and MaskRCNN improved the scalability over MLPerf-v0.6 with additional optimizations on the model (e.g., changing gather/scatter to einsum) as well as on the system (e.g., mixing SPMD partitioning with weight update sharding). The scaling is limited by communication overhead introduced for partitioning and inefficiencies from smaller dimensions after partitioning, such as spatial dimensions of later layers. The transformer model also achieves comparable speedup of 2.3$\times$ on four TPU-v3 cores. Speedup is limited by significant communication overheads in the all-reduce calls.

\begin{figure}[t]
\begin{center}
\includegraphics[width=1\columnwidth]{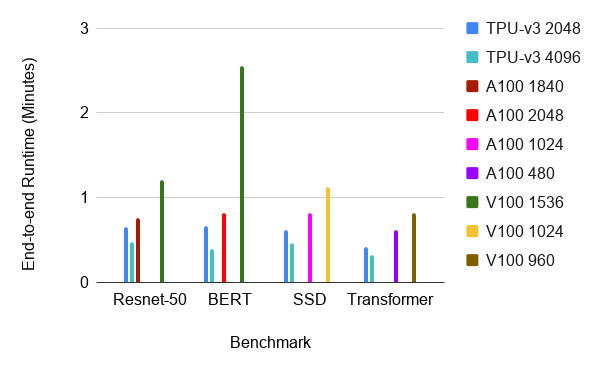}
\caption{MLPerf-v0.7 end-to-end times in minutes.}
\vspace{-1em}
\label{fig:e2etime}
\end{center}
\end{figure}

\begin{figure}[t]
\begin{center}
\includegraphics[width=1\columnwidth]{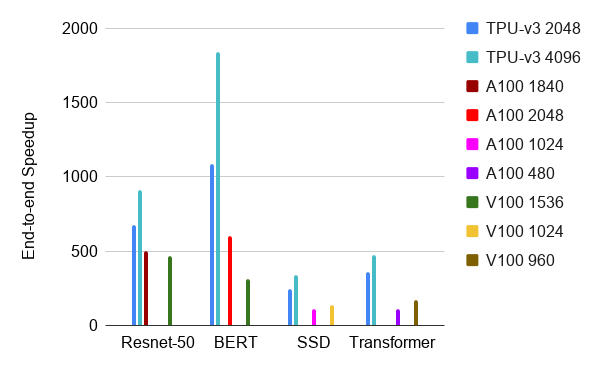}
\caption{End-to-end time speedups of MLPerf-v0.7 benchmarks over 16 accelerator chips of their own types.}
\label{fig:e2espeedup}
\end{center}
\end{figure}

\Fig{fig:e2etime} compares MLPerf benchmark end-to-end time using the TPU-v3 Multipod and NVIDIA Volta V100 and Ampere A100 GPU systems, reported by Google and NVIDIA to MLPerf-v0.7, respectively. The TPU results are in the ``Research/Development/Internal'' (RDI) category, while the GPU results are in the ``Available On-prem'' category. To compare the scalability of TPU and GPU systems,~\Fig{fig:e2espeedup} shows the topline end-to-end time speedups in MLPerf-v0.7 benchmarks over 16 accelerator chips of their own types (TPU-v3 chips or GPUs). The techniques presented in this paper enable TPUs to achieve lower end-to-end times and higher speedups.

%% file: summary.tex
\section{Conclusion and Discussion}
In this paper, we scaled ML models to the 4k-chip Google TPU-v3 Multipod machine. We invented and refined a number of techniques to achieve the best scale in six MLPerf submissions. The best parallelization varied: data parallelism in the BERT and Resnet-50 models; and model parallelism in the MLPerf SSD, MaskRCNN and Transformer models enabled training at the largest scale. A combination of aggressive compiler optimization of model parallelism, fast gradient summation at scale on the Multipod mesh, distributed evaluation, input pipeline optimizations, and model-specific optimizations contributed to reaching the highest scale. We demonstrated performance in both the TensorFlow and JAX programming frameworks. JAX’s multi-client approach further reduced startup and compilation overheads. We view the current competition in language understanding as a modern-day Space Race, with competing organizations assembling both giant machines and giant models in the quest for an Artificial General Intelligence breakthrough. The techniques in this paper are general and can be applied to other models, frameworks, and ML accelerator architectures.

We distill the lessons learnt through this cross-stack optimization experience for large-scale ML machines.
Firstly, since this machine connects 4096 nodes, a fast all-reduce communication primitive is needed. This can be challenging at a large scale, especially with model parallelism, when enabling the distributed optimizer.
Secondly, parallelism mechanisms need to be chosen based on optimizers and models. For example, we showcased data parallelism for ResNet-50 and BERT with large batch optimizers (LARS and LAMB) and a combination of efficient data and model parallelism for SSD, MaskRCNN and Transformer. It needs to be coupled with extensive hyperparameter tuning to optimize the time to convergence.
Thirdly, programming frameworks perform the best for different models. We explore and compare both JAX and TensorFlow and each framework excels at different models for higher efficiency and scale. For example, JAX has the best MLPerf-v0.7 results in two benchmarks.
Finally, each model has different scalability issues that need to be addressed. For example, the data shuffling optimization for BERT and model parallelism challenges in SSD in Section 4.

%% file: ack.tex
\section{Acknowledgement}
The authors would like to thank
Amit Sabne,
Benjamin Lee,
Berkin Ilbeyi,
Bramandia Ramadhana,
Ce Zheng,
Chi Chen,
Chiachen Chou,
David Majnemer,
David Chen,
Dimitris Vardoulakis,
Haoyu Zhang,
George Kurian,
Jay Shi,
Jeff Dean,
Jinliang Wei,
Jose Baiocchi Paredes,
Manasi Joshi,
Marcello Maggioni,
Peter Gavin,
Peter Hawkins,
Peter Mattson,
Qing Yi,
Xiangyu Dong,
Yuechao Pan,
and
Yunxing Dai
for their support and feedback.